\documentclass[letterpaper, 10 pt, journal, twoside]{ieeetran}

\IEEEoverridecommandlockouts                              % This command is only needed if 
\usepackage{graphics} % for pdf, bitmapped graphics files
\usepackage{epsfig} % for postscript graphics files
\usepackage{times} % assumes new font selection scheme installed
\usepackage{amsmath} % assumes amsmath package installed
\usepackage{amssymb}  % assumes amsmath package installed
\usepackage{subfiles} % organize into subfiles
\usepackage{blindtext} % organize into subfiles
\usepackage{multirow} % merging cells in table
\usepackage{caption} % TO-DO: double-check if we can use this package
\usepackage{tabularx}
\usepackage{subcaption} % subfigures 
\usepackage{mathtools, nccmath} % custom rounding notation
\usepackage{booktabs} % toprule/bottom rule
\usepackage[dvipsnames]{xcolor} % text color
\usepackage{soul} % hl
\usepackage{comment}
\usepackage{placeins} % force figures to all appear before bib (float barrier)
\usepackage[para]{threeparttable}

\makeatletter
\let\NAT@parse\undefined
\makeatother
\makeatletter
\DeclareRobustCommand{\iscircle}{\mathord{\mathpalette\is@circle\relax}}
\newcommand\is@circle[2]{%
  \begingroup
  \sbox\z@{\raisebox{\depth}{$\m@th#1\bigcirc$}}%
  \sbox\tw@{$#1\square$}%
  \resizebox{!}{\ht\tw@}{\usebox{\z@}}%
  \endgroup
}
\makeatother

\usepackage{hyperref}
\title{
DecTrain: Deciding When to Train a Monocular Depth DNN Online
}
\author{Zih-Sing Fu$^*$, Soumya Sudhakar$^*$, Sertac Karaman, Vivienne Sze % <-this % stops a space
\thanks{Manuscript received: August, 15th, 2024; Revised November, 24, 2024; Accepted January, 15, 2025.}
\thanks{This paper was recommended for publication by Editor Abhinav Valada upon evaluation of the Associate Editor and Reviewers' comments.}
\thanks{$^*$The first two authors contributed equally to this work. Authors are with the Massachusetts Institute of Technology, Cambridge, MA  02139,  USA.  Emails: {\tt\{zihsing, soumyas, sertac, sze\}@mit.edu}. This work was funded by National Science Foundation Real-Time Machine
Learning program grant no. 1937501, MIT Mobility Initiative, MIT Accenture Fellowship, Huang Phillips Fellowship, MathWorks Fellowship, and
Intel.}
\thanks{Digital Object Identifier (DOI): 10.1109/LRA.2025.3536206.}% <-this % stops a space
}

\begin{document}

\maketitle
%\thispagestyle{empty}
%\pagestyle{empty}
% added for final

\IEEEpubid{\begin{minipage}{\textwidth} \centering
    \vspace{48pt}
    {\footnotesize © 2025 IEEE. Personal use of this material is permitted. Permission from IEEE must be obtained for all other uses, in any current or future media, including reprinting/republishing this material for advertising or promotional purposes, creating new collective works, for resale or redistribution to servers or lists, or reuse of any copyrighted component of this work in other works.}
\end{minipage}}
%%%%%%%%%%%%%%%%%%%%%%%%%%%%%%%%%%%%%%%%%%%%%%%%%%%%%%%%%%%%%%%%%%%%%%%%%%%%%%%%
\begin{abstract} Deep neural networks (DNNs) can deteriorate in accuracy when deployment data differs from training data. While performing online training at all timesteps can improve accuracy, it is computationally expensive. We propose DecTrain, a new algorithm that decides when to train a monocular depth DNN online using self-supervision with low overhead. To make the decision at each timestep, DecTrain 
% predicts the accuracy gain from training and compares it to the cost of training.
compares the cost of training with the predicted accuracy gain.
% to decide when to perform online training. 
We evaluate DecTrain on out-of-distribution data, and find DecTrain maintains accuracy compared to online training at all timesteps, while training only 44\% of the time on average. We also compare the recovery of a low inference cost DNN using DecTrain and a more generalizable high inference cost DNN on various sequences. DecTrain recovers the majority (97\%) of the accuracy gain of online training at all timesteps while reducing computation compared to the high inference cost DNN which recovers only 66\%. With an even smaller DNN, we achieve 89\% recovery while reducing computation by 56\%. DecTrain enables low-cost online training for a smaller DNN to have competitive accuracy with a larger, more generalizable DNN at a lower overall computational cost.
\end{abstract}


%%%%%%%%%%%%%%%%%%%%%%%%%%%%%%%%%%%%%%%%%%%%%%%%%%%%%%%%%%%%%%%%%%%%%%%%%%%%%%%%
\section{Introduction}
\par \IEEEPARstart{C}{onsider} the case in which a resource-constrained robot equipped with a single RGB camera and IMU navigates in an environment using a deep neural network (DNN) to predict depth per pixel of the RGB image. While using a DNN for the task of monocular depth estimation can be more energy efficient and has a smaller form factor than traditional bulky and high-power physical depth sensors such as LiDAR or active IR stereo~\cite{intelrealsense, velodyne}, DNNs are prone to accuracy degradation on images that differ from those of the training distribution in a range of domains~\cite{kendall2017uncertainties, amini2020deep, blum2019fishyscapes, koh2021wilds}. 
In monocular depth estimation, out-of-distribution accuracy degradation can occur during deployment when the objects are novel or the scale of depth changes drastically. 
Accuracy degradation in out-of-distribution environments motivates the need for adaptation of the DNN to its deployment environment via online training, which has been shown to meaningfully improve accuracy when deployed~\cite{vodisch2023codeps}.
\begin{figure}[t]
    \centering
    \includegraphics[width=1.\columnwidth]{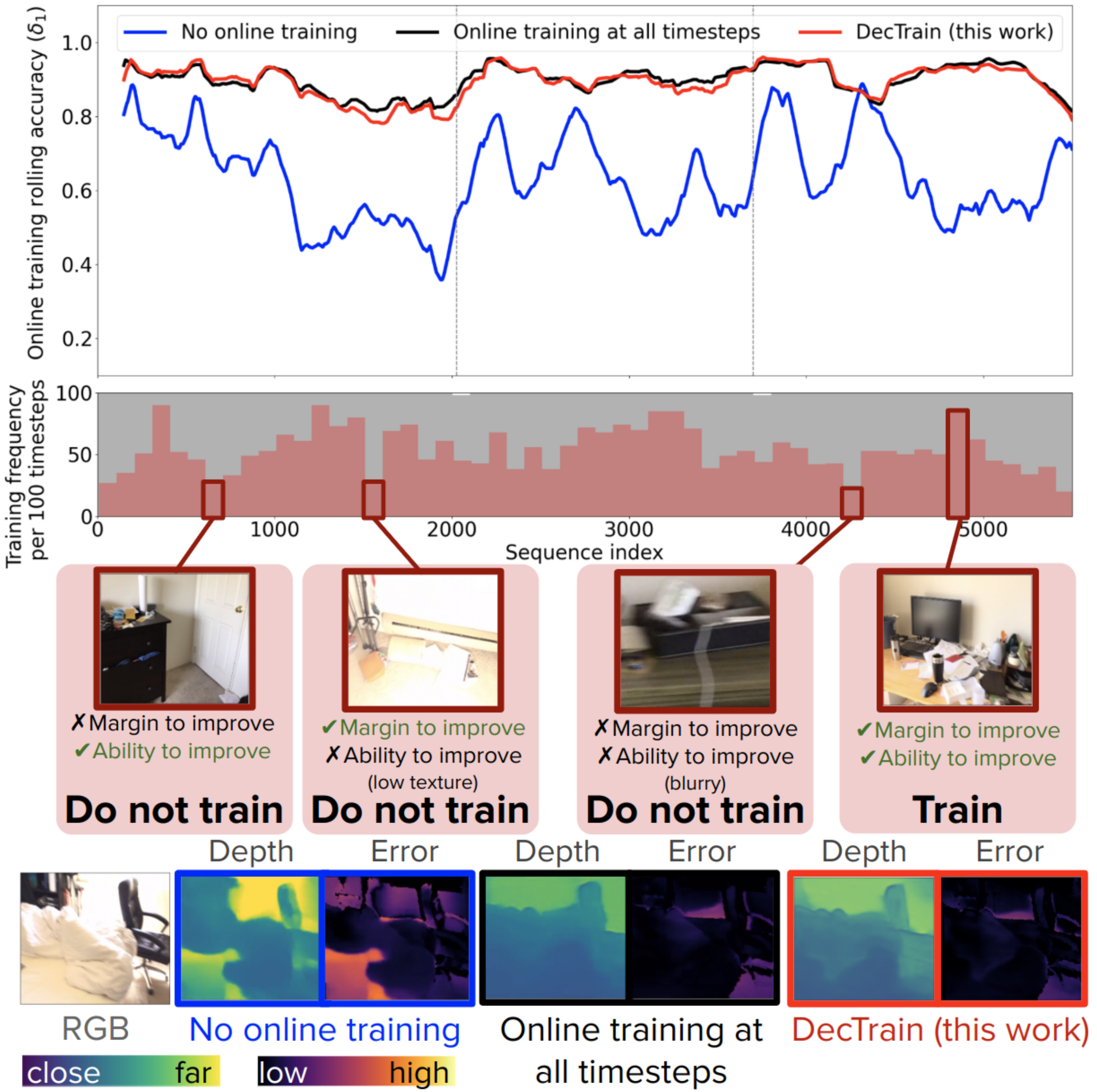}
    \caption{DecTrain (red) decides when to perform online training based on \textit{margin to improve} (visualized by the gap between the blue and black lines) and \textit{ability to improve} (visualized by the texture and sharpness in the image). Compared to the baseline of online training at all timesteps (black) or no timesteps (blue), DecTrain maintains the accuracy improvement of adaptation while training on only a subset of the timesteps (dashed lines denote new sequence). 
    }
    \label{fig:on_the_fly_learning_teaser}
    \vspace{-12pt}
\end{figure}
\par We consider the case where the training is \emph{online} and \emph{onboard} the robot. Online training means that (1) the robot receives images sequentially from a camera video stream from which each image and pose pair $(X_{t},\mathcal{P}_t)$ is obtained one-by-one at timestep $t$ where $t=[t_{1},...,t_{N}]$ and $N$ is the total number of timesteps in the camera trajectory, and (2) the robot requires self-supervision to learn since there is no access to ground-truth depth. Since the online training also occurs onboard the resource-constrained platform, it is critical to limit the computational cost of online training to reduce energy consumption. 
\par The computational cost for inference and performing online training for a DNN can be divided into four components: (1) the number of timesteps we run inference ($N_{inf}$), (2) the cost per inference ($C_{inf}$), (3) the number of timesteps we run training ($N_{train})$, and (4) cost per training update ($C_{train})$. The total cost of monocular depth estimation ($C_{tot}$) is     
\begin{equation}
    C_{tot} = N_{inf} \cdot C_{inf} + N_{train} \cdot C_{train}.
\end{equation}
The robot runs an inference to predict depth at every timestep ($N_{inf} = N$) when it receives a new image and pose pair $(X_{t},\mathcal{P}_t)$.
Unlike previous works where either no online training is performed ($N_{train} = 0$)~\cite{ranftl2021vision, bhat2023zoedepth, oquab2023dinov2} as seen by the blue curve in Fig.~\ref{fig:on_the_fly_learning_teaser}
or online training is performed at all timesteps to improve accuracy ($N_{train} = N$)~\cite{vodisch2023codeps} as seen by the black curve in Fig.~\ref{fig:on_the_fly_learning_teaser}, we study the case where the robot decides whether or not to train at each timestep such that $0 \leq N_{train} \leq N$.
Any additional overhead to make a decision ($C_{decision}$) also adds to the total compute cost of monocular depth estimation such that
\begin{equation}
    C_{tot} = N \cdot C_{inf} + N_{train} \cdot C_{train} + N \cdot C_{decision}.
    \label{eq:compute_cost_formula}
\end{equation}
Clearly, reducing $N_{train}$ can decrease the total cost of depth estimation compared to online training at all timesteps if the overhead does not outweigh the savings,
\textit{i.e.,}
\begin{equation}
    N \cdot C_{decision} < (N-N_{train}) \cdot C_{train}.
\end{equation}
 
\par Our main contribution is a new method called DecTrain that decides whether to train a monocular depth DNN at each timestep based on when the potential accuracy improvement is worth the computational cost of training.
Compared to performing online training at all timesteps, DecTrain is able to reduce the percentage of timesteps we train to 30-58\% while maintaining accuracy on 10 representative experiments.
We also show that low inference cost DNNs using DecTrain can achieve competitive accuracy (4-6\% higher accuracy) and lower computational cost (17-57\% lower GFLOPs) compared to a high inference cost state-of-the-art DNN across 100 out-of-distribution sequences.

%%%%%%%%%%%%%%%%%%%%%%%%%%%%%%%%%%%%%%%
\section{Related Work}
\subsection{Domain adaptation} 
Recent research has focused on unsupervised domain adaptation, where the DNN is trained online to perform better on the current environment; when the goal is also to preserve accuracy on previous environments, the algorithms are called continual or lifelong learning~\cite{vodisch2023codeps}. Continual learning algorithms often involve replay buffers~\cite{kuznietsov2021comoda}, regularization, and increasing architecture capacity~\cite{parisi2019continual}. 
Similar to the CoDEPS framework by Vodisch et al.~\cite{vodisch2023codeps} that continuously trains a DNN online to improve performance on the deployment environment, our problem involves online training a monocular depth DNN;
unlike previous works, we aim to reduce the frequency of online training rather than perform online training at all timesteps.  
In addition, our priority is to maximize accuracy on the current environment rather than accept a lower accuracy on the current environment to preserve accuracy on the pretraining environment. 
\subsection{Sample selection during training} 
Previous work to save computation in the offline training setting used a linearized model of training dynamics to cut-off training early~\cite{zancato2020predicting}. In contrast, in the online setting, training dynamics are not easy to model, motivating a data-driven approach. Selecting an informative subset of samples has also been studied in the active learning community. 
Given a DNN that has been trained on a dataset $\nu$, the traditional active learning set-up is to select a subset of data $\nu_{\text{new\_labels}}$ from an unlabeled dataset to be labeled that will maximize accuracy after retraining the DNN on $\nu \cup \nu_{\text{new\_labels}}$~\cite{settles2009active}. 
Unlike previous works, our motivation is to reduce training effort instead of reducing labeling effort.  
\par The majority of research focus has been on the offline pool-based scenario, where the entire unlabeled dataset is accessible when making a decision, the ground-truth labels are available,
and the DNN can be retrained on $\nu \cup \nu_{\text{labeled}}$ ~\cite{settles2009active} for tasks ranging from image classification~\cite{gal2017deep} to semantic segmentation~\cite{siddiqui2020viewal}. 
Work where data is revealed sequentially still assumes ground-truth labels are available which is not the case in an online setting~\cite{narr2016stream}. 
Determining whether to select a piece of data to label in prior works has been based on uncertainty~\cite{gal2017deep}, 
diversity~\cite{sener2018active},
and hybrid combinations of both~\cite{yang2015multi}. 
In this work, we incorporate DNN uncertainty and other relevant variables to make a decision.

\subsection{DNN uncertainty estimation}
There are two types of DNN uncertainty to measure: (1) aleatoric uncertainty (noise inherent in the data that is irreducible by additional training), which can be computed relatively inexpensively via a modified loss function~\cite{kendall2017uncertainties},
and (2) epistemic uncertainty (noise inherent to the model reducible by additional training), which is commonly calculated using computationally expensive approaches such as Monte-Carlo Dropout (MC-Dropout)~\cite{gal2016dropout}, Bayesian neural networks~\cite{blundell2015weight}, or ensembles~\cite{lakshminarayanan2017simple}, leading to recent works looking into reducing the cost using evidential learning~\cite{amini2020deep} or temporal redundancy~\cite{sudhakar2022uncertainty}. Here, we utilize both aleatoric and epistemic uncertainty to make a decision on when to train and select a computationally efficient uncertainty estimation method~\cite{sudhakar2022uncertainty} to limit the overhead of the decision.

\subsection{Markov decision processes for modeling decision-making} 
Markov decision processes (MDPs) have been used in metareasoning algorithms for motion planning that decide when to stop computing~\cite{sung2021learning}
or when to tune hyperparameters~\cite{bhatia2022tuning}.
We also cast our problem of deciding whether to train at each timestep as a MDP, and showcase a greedy approach to learn the one-step reward of training.
%%%%%%%%%%%%%%%%%%%%%%%%%%%%%%%%%%%%%%%
\section{Problem Definition}
\label{subsec:problem_setup}
\begin{figure*}[t!] \centering
     \centering
     \includegraphics[width=\textwidth]{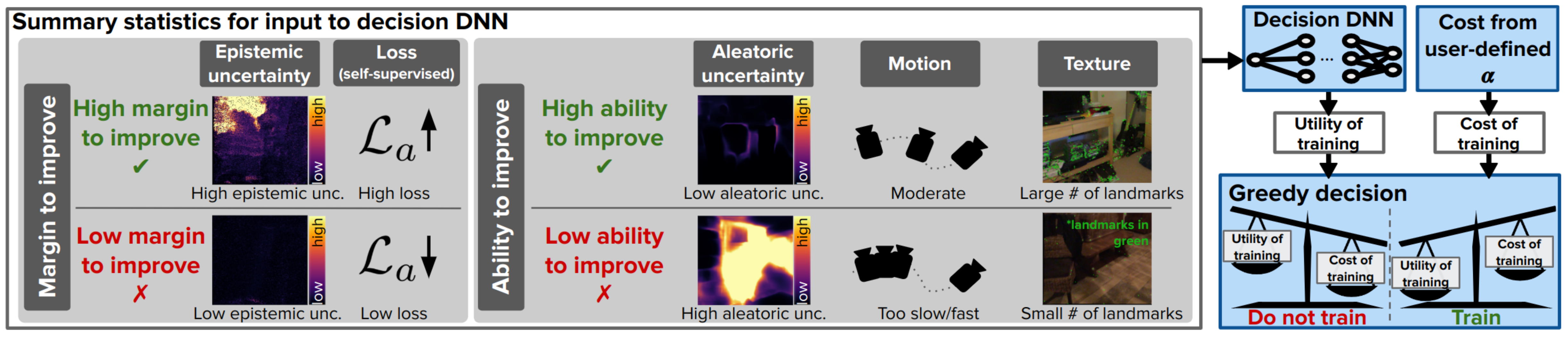}
     \caption{DecTrain overview: at each timestep, the decision DNN takes inputs relevant to the margin and ability to improve to predict the utility of training, which is compared to the cost of training to decide when to train the monocular depth DNN.}
     \label{fig:diagram}
     % \vspace{-0.5cm}
\end{figure*}
We study the problem of deciding when to perform online training to reduce the cost of adaptation for a monocular depth DNN. 
We formulate the problem as a MDP $\mathcal{M} = \langle \mathcal{S, A, T, R}, \gamma, s_{0}\rangle$ defined by the state space $s_{t} \in \mathcal{S}$, the action space $a_{t} \in \mathcal{A}$, the transition function $\mathcal{T}(s_{t+1}|s_{t}, a_{t})$, the reward function $R(s_{t}, a_{t}, s_{t+1})$, the discount factor $\gamma$, and the initial state $s_{0}$. 
Here, the state at time $t$ is $s_{t} = [\mathcal{L}(X_{t-w:t}, D(X_{t},\theta_{k})), \psi_{t}]$ which is made up of (1) the self-supervised loss $\mathcal{L}(X_{t-w:t}, D(X_{t},\theta_{k}))$ which is dependent on the current image and previous $w$ images $X_{t-w:t}$ and the depth prediction $D(X_{t},\theta_{k})$ at time $t$ which itself is dependent on the current monocular depth DNN $\theta_{k}$ and current image $X_{t}$ , and (2) a feature vector $\psi_{t}$ that represents aspects of the robot and DNN relevant to the decision. 
\par The two actions available at each timestep are $a_{t} \in \{T, \overline{T}\}$ to train ($T$) or not train ($\overline{T}$) the DNN $\theta_{k}$. Training the DNN means we run an online training process $\mathcal{G}$ to obtain an updated set of DNN weights $\theta_{k} \rightarrow \theta_{k+1}$ for the next timestep. 
Not training the DNN means we still have the current set of weights $\theta_{k}$ for the next timestep. 
We choose an existing online training process with a self-supervised loss based on the current and previous two images ($w=2$)~\cite{vodisch2023codeps}.
\par Importantly, the transition function $\mathcal{T}(s_{t+1}|s_{t}, a_{t})$ is unknown since (1) we do not know what the new weights of DNN $\theta_{k+1}$ will be before we perform the online training process $\mathcal{G}$, (2) we do not have access to the next image $X_{t+1}$ in the online setting. 
The reward function is 
\begin{equation} \label{eq:reward_func}
    R_{t} = \begin{cases}
        \frac{-1}{\alpha} + \mathcal{U}(s_{t}, a_{t}=T, s_{t+1}) & \text{if } a_{t} = T \\ 
        0 & \text{if } a_{t} = \overline{T},  \\ 
    \end{cases}
\end{equation}
where $ \frac{-1}{\alpha}$ represents the cost of training with a user-defined $\alpha$ to balance training cost with potential accuracy improvement, $\mathcal{U}(s_{t}, a_{t}=T, s_{t+1})$ represents the utility of training, and $R_{t} = R(s_{t}, a_{t}, s_{t+1})$ for brevity. 
$\mathcal{U}(s_{t}, a_{t}=T, s_{t+1})$ is unknown in most cases since we do not know the transition function.
Our reward formulation exposes the design decision to the user on how to balance between the cost of training and accuracy improvement via $\alpha$; for a resource-constrained robot robust to collisions, 
the user may select a low $\alpha$ to only spend the compute to train when the potential accuracy improvement is high, while for a less resource-constrained robot fragile to collisions,
the user may select a high $\alpha$ to train even when the potential accuracy improvement is low. 
\par The optimal solution to the MDP is the policy that maximizes the discounted reward such that
\begin{equation}
    \pi^*(s_{t}) = \text{argmax}_{\pi} \sum_t^N \gamma^{t-1}R_{t}
\end{equation}
for all states. Since the transition function is unknown, we propose an approximate solution described in Section~\ref{sec:dectrain}.

%%%%%%%%%%%%%%%%%%%%%%%%%%%%%%%%%%%%%%%
\section{DecTrain: Deciding When to Train}
\label{sec:dectrain}
We propose DecTrain, a method that decides whether to train at each timestep based on whether the predicted utility of training is worth the cost of training. 
DecTrain approximates a solution to the MDP formulation in Section~\ref{subsec:problem_setup} by
\begin{enumerate}
    \item predicting the utility of training using a compute-efficient decision DNN that takes inputs relevant to the margin to improve and the ability to improve, and
    \item greedily deciding to train if the reward for training is larger than the reward for not training.
\end{enumerate}
To approximately solve the MDP, we need to set the utility of training function $\mathcal{U}(s_{t}, a_{t}=T, s_{t+1})$.
We set it to be the relative change in self-supervised loss after training such that 
\begin{equation} 
    \begin{split}
        & \mathcal{U}(s_{t}, a_{t}=T, s_{t+1}) = (\mathcal{L}_{a}-\mathcal{L}_{b})/\mathcal{L}_{a},\\
    \end{split}
\end{equation}
where for brevity, $\mathcal{L}_{a} = \mathcal{L}(X_{t-1:t+1}, D(X_{t+1}, \theta_{k}))$ is the loss on the next image before we train the DNN and $\mathcal{L}_{b}=\mathcal{L}(X_{t-1:t+1}, D(X_{t+1}, \theta_{k+1}))$ is the loss on the next image after we train the DNN. Note, there is a sign change since positive changing utility is decreasing self-supervised loss. This definition of utility makes $\alpha$ intuitive to choose, where it is worth training if the loss can be reduced by over $\frac{1}{\alpha}$. 
For example, at $\alpha=500$, DecTrain should decide to train if it predicts the loss can be reduced by $\frac{1}{\alpha}\times100\% = 0.2\%$.
We next step through the two components of DecTrain: predicting the utility of training and the greedy policy. 
\subsection{Learning to predict the utility of training}
\label{subsec:utility_prediction}
When DecTrain is making a decision, both $\mathcal{L}_{a}$ and $\mathcal{L}_{b}$ are unknown 
since we have no access to the next camera image $X_{t+1}$.  
Since we have no way to access future camera images, we will assume no motion, such that $X_{t+1} = X_{t}$, such that
\begin{equation}
\begin{split}
    \mathcal{U}&(s_{t}, a_{t}=T, s_{t+1})\approx(\widetilde{\mathcal{L}_{a}}-\widetilde{\mathcal{L}_{b}})/\widetilde{\mathcal{L}_{a}}, \\
\end{split}
\label{eq:utility_dectrain}
\end{equation}
where for brevity, we refer to the loss on the \textit{current} image before training as $\widetilde{\mathcal{L}_{a}}=\mathcal{L}(X_{t-2:t}, D(X_{t},\theta_{k}))$
and the loss on the \textit{current} image after training as $\widetilde{\mathcal{L}_{b}}=\mathcal{L}(X_{t-2:t}, D(X_{t}, \theta_{k+1}))$.
While $\widetilde{\mathcal{L}_{a}}$ is known when DecTrain must make a decision, $\widetilde{\mathcal{L}_{b}}$ is unknown since the updated DNN $\theta_{k+1}$ remains unknown before we perform online training. 
To predict the utility $\mathcal{U}(s_{t}, a_{t}=T, s_{t+1})$, we propose using a decision DNN $\phi_{k}$ that predicts an approximate utility $\widetilde{\mathcal{U}}(s_{t}, a_{t}=T, s_{t+1})$ given an informative $42\times1$ input vector $\psi_{t}$ relevant to \textit{the margin to improve} and \textit{the ability to improve} at timestep $t$ as seen in Fig.~\ref{fig:diagram}. 
\subsubsection{Inputs relevant to margin to improve}  
We refer to the margin to improve to be the gap in accuracy between how well the monocular depth DNN is currently performing and how well it could perform with perfect guidance. For example, a depth prediction with 40\% accuracy has more margin to improve than a depth prediction with 95\% accuracy. However, given we have no access to ground-truth depth in the online setting, we cannot compute the actual accuracy. Instead, DecTrain uses the epistemic uncertainty of the depth prediction (capturing the uncertainty in the model weights that can be reduced by training) and the current loss to indicate low or high margin to improve as seen in Fig.~\ref{fig:diagram}.
\subsubsection{Inputs relevant to ability to improve}
\par Online training uses self-supervision, which provides noisy guidance, requiring us to consider the \textit{ability to improve}. We refer to the ability to improve as the ability of the self-supervision to guide the DNN weights in the correct direction to improve accuracy. For example, photometric error, which is part of the off-the-shelf self-supervised loss we use~\cite{vodisch2023codeps}, can fail in cases of low image texture or when the motion is too fast or too slow. 
In addition, there are sources of error that cannot be reduced with online training due to missing data in the image (\textit{e.g.}, a completely dark patch of an image). To capture the ability to improve, we include the aleatoric uncertainty of the depth prediction (capturing the uncertainty in the data that will not be reduced by training), the poses of the robot in the current and previous two frames (capturing motion), and the number of landmarks extracted using SIFT~\cite{lowe2004distinctive} in the current and previous two frames (capturing texture) to indicate low or high ability to improve as seen in Fig.~\ref{fig:diagram}.
\par To keep $\psi_{t}$ compact for reducing the memory and compute overhead, we only use summary statistics (mean, median, maximum, and minimum) for the above quantities. 

\subsubsection{Training decision DNN} 
We can train $\phi_{k}$ both offline using collected data on $\mathcal{U}(s_{t},a_{t}=T,s_{t+1})$ and $\psi_{t}$, and online where after online training the monocular depth DNN $\theta_{k} \rightarrow \theta_{k+1}$, we can run one additional depth inference on the current image $X_{t}$ with $\theta_{k+1}$ to get the ground-truth $\mathcal{U}(s_{t},a_{t}=T,s_{t+1})$ and train the decision DNN $\phi_{k}\rightarrow \phi_{k+1}$.

\subsection{Greedy decision-making}
Reasoning about future states is challenging since future states depend on future camera measurements $X_{t+i}$ for $i = 1, 2, ..., \infty$ which are completely unknown and not controllable by our action space ($a_{t} \in \{T, \overline{T}\}$). Therefore, we assume no motion to approximate the utility function in Section~\ref{subsec:utility_prediction}. Since this assumption is only valid for the near horizon given the robot is moving and the environment can be dynamic, we greedily decide whether to train at the current timestep $t$ based on the reward at $t$ ($\gamma = 0$) as in 
\begin{equation}
\begin{split}
        \pi(s_{t}) & = \text{argmax}_{a_{t}} R(s_{t}, a_{t}, s_{t+1}) \\
        & = \begin{cases}
         a_{t} = T & \text{if } \frac{-1}{\alpha} + \widetilde{\mathcal{U}}(s_{t}, a_{t}=T, s_{t+1}) > 0, \\ 
        a_{t} = \overline{T} & \text{else.}   \\ 
    \end{cases}
    \label{eq:dectrain-decision}
\end{split}
\end{equation}
Since it is computationally intractable to brute-force a non-myopic oracle, we compare performance of DecTrain to a greedy oracle that uses the same policy in Eq.~\ref{eq:dectrain-decision}, but with the ground-truth utility $\mathcal{U}(s_{t}, a_{t}=T, s_{t+1})$. 

\par In summary, at each timestep, DecTrain predicts the utility of training
and makes a greedy decision whether or not to perform online training based on balancing the cost of training with the predicted utility of training. Code will be made available at \url{https://lean.mit.edu/papers/dectrain}.
We next evaluate DecTrain and show it is able to improve accuracy while reducing the computation.

%%%%%%%%%%%%%%%%%%%%%%%%%%%%%%%%%%%%%%%
\section{Experimental Setup}
\subsection{Pretraining monocular depth DNN and decision DNN}

\subsubsection*{Monocular depth DNN ($\theta_{0}$)} We follow the model architecture design in previous work~\cite{vodisch2023codeps}.
In detail, we use a ResNet-101~\cite{he2016deep} shared encoder, followed by (1) a Monodepth2~\cite{godard2019digging} depth decoder 
and (2) a Monodepth2 aleatoric uncertainty decoder.
We pretrain 5 models with randomly initialized weights on the NYUDepthV2~\cite{silberman2012indoor} dataset with downsampled resolution 224$\times$224, negative log-likelihood loss~\cite{kendall2017uncertainties}, and an Adam optimizer with learning rate of 0.0001. 
The pre-trained models achieve 83.9\% $\delta_1$ validation accuracy on NYUDepthV2. To evaluate lower-cost DNNs, we also train DNNs with ResNet-18 and ResNet-50 encoders. 

\subsubsection*{Decision DNN ($\phi_{0}$)} We use a 3-layered fully connected DNN with hidden dimensions set to 32.
For pretraining, we collect the inputs and the ground-truth utility on 5 ScanNet~\cite{dai2017scannet} sequences not used in testing, use a loss made up of the mean squared error and correlation coefficient, and an Adam optimizer with learning rate of 0.001.

\subsection{Online training process}
\subsubsection*{Monocular depth DNN ($\theta_{k}\rightarrow\theta_{k+1}$)}
If DecTrain decides to train, we follow the online training process used in previous work~\cite{vodisch2023codeps}.
In detail, the depth decoder is trained using a self-supervised loss based on the photometric error, structural similarity, and edge-aware smoothness~\cite{vodisch2023codeps}.
We use provided pose estimates~\cite{dai2017scannet, xiao2013sun3d, liao2022kitti} and an Adam optimizer with learning rate of 0.0001 on a batch
of the current image and two randomly replayed images from a replay buffer.
The replay buffer stores data on up to 300 timesteps in the current environment. 
Following previous work~\cite{vodisch2023codeps}, the replay buffer is updated when the current image differs significantly from the images stored.
We freeze the encoder following previous work~\cite{vodisch2023codeps}, and we freeze the aleatoric uncertainty decoder to maintain stability during online training. 

\subsubsection*{Decision DNN ($\phi_{k}\rightarrow\phi_{k+1}$)} The decision DNN is trained if DecTrain decides to train.
After updating the depth decoder,
we run an additional depth decoder inference to get the label of ground-truth utility.
We use the same loss and optimizer as in pretraining for 10 epochs with data from the current timestep and 32 randomly selected timesteps from both the target replay buffer and the source replay buffer for the decision DNN. 
The decision DNN target replay buffer is a FIFO buffer storing up to 300 inputs and ground-truth utility that gets added to when DecTrain decides to train. The decision DNN source replay buffer stores 3000 randomly selected data from the pretraining dataset. 
We include data from the pretraining dataset since we only get a label online if the train action is taken, leading to a biased label collection.

To measure the computational cost, we use NVIDIA PyProf~\cite{nvidia-pyprof} and PyPAPI~\cite{pypapi} to obtain the average number of giga floating-point operations (GFLOPs) per timestep.
%%%%%%%%%%%%%%%%%%%%%%%%%%%%%%%%%%%%%%%
\section{Experimental Results}
\label{sec:exp}
% Section: Experiment Results
In this section, we evaluate DecTrain on 133 sequences from out-of-distribution indoor and outdoor datasets (ScanNet~\cite{dai2017scannet}, SUN3D~\cite{xiao2013sun3d}, KITTI-360~\cite{liao2022kitti}) as seen in Table~\ref{table:datasets}. 
\begin{table}[t]
\vspace{0.15cm}
\begin{center}
\caption{Experiments on multiple sequences in single environments (Exp. 1-4), multiple environments (Exp. 5, 11), and single sequences in single environments (Exp. 6-10, 12).
}
\begin{tabular}{c c c c}
\toprule
 Exp. & Dataset & Online training & Validation \\
 \toprule
 1 & ScanNet & 0101\_01, 0101\_02, 0101\_03, 0101\_04 & 0101\_05\\
 2 & ScanNet & 0673\_01, 0673\_02, 0673\_03, 0673\_04 & 0673\_05\\
 3 & ScanNet & 0451\_00, 0451\_02, 0451\_03, 0451\_04 & 0451\_05\\
 4 & ScanNet & 0092\_01, 0092\_02, 0092\_03 & 0092\_04\\
 5 & ScanNet & 0101\_01, 0101\_02, 0101\_03, 0101\_04, & 0101\_05\\
   &         & 0673\_01, 0673\_02, 0673\_03, 0673\_04, & 0673\_05\\
   &         & 0451\_00, 0451\_02, 0451\_03, 0451\_04, & 0451\_05\\
   &         & 0092\_01, 0092\_02, 0092\_03 & 0092\_04 \\
6 & SUN3D & d6\_lounge\_1, first 70\% & last 30\%\\
7 & SUN3D & tian\_lab\_1, first 70\% & last 30\%\\
8 & SUN3D & playroom\_1, first 70\% & last 30\%\\
9 & SUN3D & 123\_1, first 70\% & last 30\%\\
10 & SUN3D & 124\_1, first 70\% & last 30\%\\
11 & KITTI-360 & 00, 01, 03, 04, 05, 06, 07, 09  & 10  \\
12 & ScanNet & 100 randomly selected sequences & --\\
\bottomrule
\end{tabular}
\label{table:datasets}
\end{center}
\end{table}
\begin{table*}[t]
\vspace{0.2cm}
\centering
\caption{Accuracy (online training accuracy ($\delta_{1}$), validation accuracy ($\delta_{1, val}$)) and computing cost
(GFLOPs) metrics
for baselines
and DecTrain with $\alpha=500$ on Exp. 1-10 averaged over five trials with
maximum standard deviation of $\delta_{1}$ of 6.4\%, 1.7\%, and 2.2\% for no online training, online training at all timesteps, and DecTrain respectively.  
We see that DecTrain can maintain accuracy within 1\% of online training at all timesteps while reducing the computation.}
\begin{minipage}{\textwidth}
\begin{tabularx}{\textwidth}{@{\extracolsep{\fill}}c |c c c | c c c |c c c c}
  \toprule 
 & \multicolumn{3}{c}{No online training} & \multicolumn{3}{c}{Online training at all timesteps} &\multicolumn{4}{c}{DecTrain} \\
 Exp. & $\delta_{1}$ (\%) & $\delta_{1, val}$ (\%) & GFLOPs & $\delta_{1}$ (\%) & $\delta_{1, val}$ (\%) & GFLOPs & $\delta_{1}$ (\%) & $\delta_{1, val}$ (\%) & \% trained & GFLOPs \\
 \midrule
 1  & 82.1 & 87.5 & 21    & 86.4 & 89.7 & 88    & 87.4 & 90.7 & 41 & 62 \\
 2  & 68.9 & 64.5 & 21    & 90.8 & 83.9 & 88    & 90.5 & 84.0 & 47 & 66 \\
 3  & 70.4 & 78.6 & 21    & 86.3 & 81.3 & 88    & 85.5 & 81.0 & 43 & 63 \\
 4  & 62.3 & 66.1 & 21    & 87.5 & 73.0 & 87    & 86.8 & 73.4 & 55 & 72 \\
 5  & 70.6 & 74.2 & 21    & 87.8 & 82.0 & 88    & 87.6 & 82.7 & 44 & 65 \\
 6  & 47.1 & 67.4 & 21    & 70.0 & 36.6 & 88    & 69.5 & 38.8 & 30 & 55 \\
 7  & 52.9 & 47.8 & 21    & 68.4 & 45.1 & 88    & 67.8 & 47.4 & 54 & 72 \\
 8  & 68.1 & 50.5 & 21    & 85.1 & 59.3 & 88    & 84.2 & 54.8 & 39 & 61 \\
 9  & 50.0 & 67.2 & 21    & 73.1 & 70.9 & 88    & 74.8 & 71.0 & 31 & 55 \\
 10 & 43.9 & 48.7 & 21    & 61.6 & 45.7 & 88    & 60.8 & 48.8 & 58 & 75 \\
 \midrule
 Avg. &  61.6 & 65.3 & 21    & 79.7 & 66.8 & 88   & 79.5 & 67.3 & 44 & 65 \\
 \bottomrule
\end{tabularx}
\label{table:comp_acc_all_sequences}
\end{minipage}
\end{table*}

\begin{figure}[t!] \centering
     \centering
     \includegraphics[width=0.9\columnwidth]{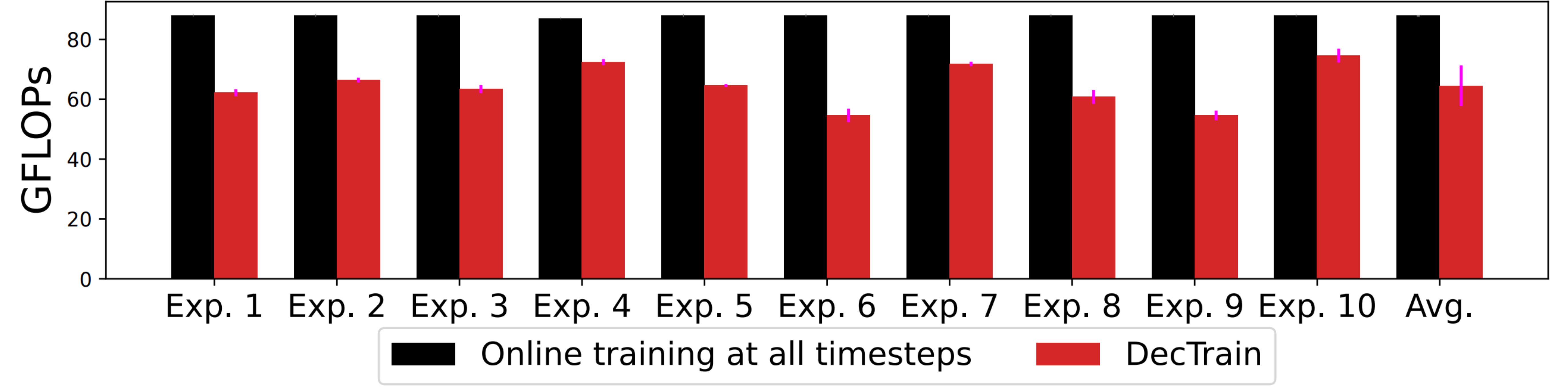}
     \caption{
     DecTrain lowers computation by 27\% vs.
     online training at all timesteps. Error bars are one standard deviation.}
     \label{fig:acc_gflops_barchart}
\end{figure}
\begin{figure*}[t!] \centering
     \centering
     \includegraphics[width=0.99\textwidth]{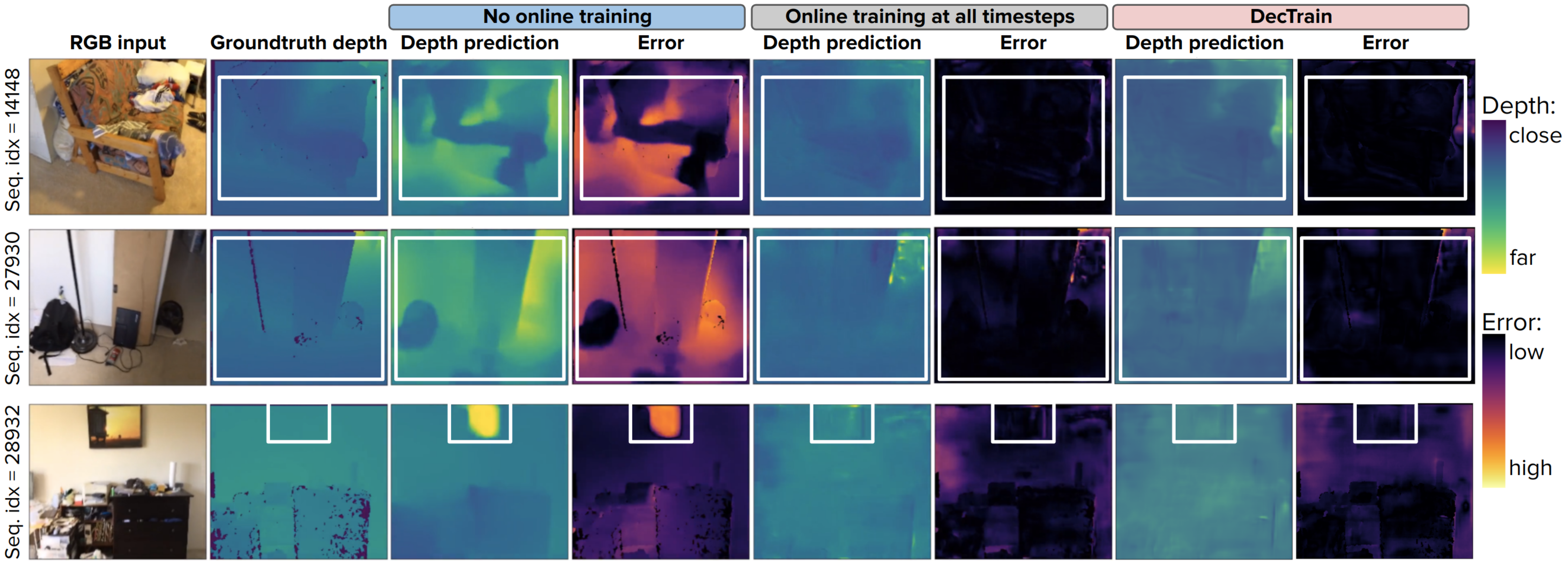}
     \caption{Examples of RGB input, ground-truth depth, depth prediction, and error for baselines and DecTrain from Exp. 5. Compared to no online training, there is lower error in the online training at all timesteps and DecTrain (see white boxed regions). Online training improves the depth scale (\textit{e.g.}, walls and floor in second row) and the prediction for unknown objects (\textit{e.g.}, painting in third row), and DecTrain mimics the improved performance at lower computational cost. While the off-the-shelf self-supervised loss~\cite{vodisch2023codeps} improves the scale, it also introduces artifacts that are visually less smooth. These artifacts are also present when using online training at all timesteps due to the loss function, not DecTrain.
     }
     \label{fig:qualitative_imgs}
\end{figure*}
\begin{figure}[t]
    \centering
    \includegraphics[width=0.9\columnwidth]{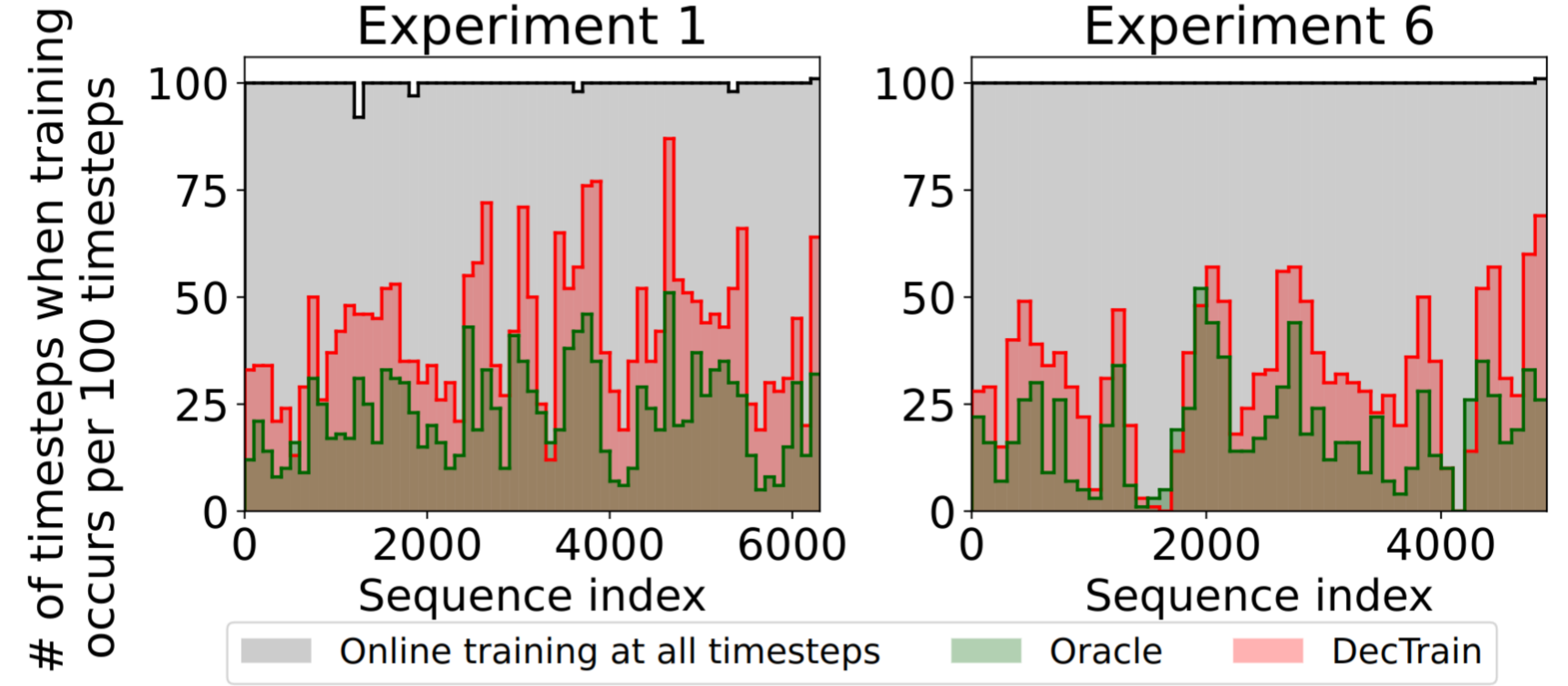}
    \caption{Histograms of training decisions on experiments from ScanNet and Sun3D. DecTrain and the greedy oracle reduce the amount of training vs. online training at all timesteps, and DecTrain closely follows
    the greedy oracle. Note, online training can only run when a SLAM pose is available.
    }
    \label{fig:on_the_fly_learning_histogram}
\end{figure}
\begin{figure}[t]
    \centering
    \includegraphics[width=1.\columnwidth]{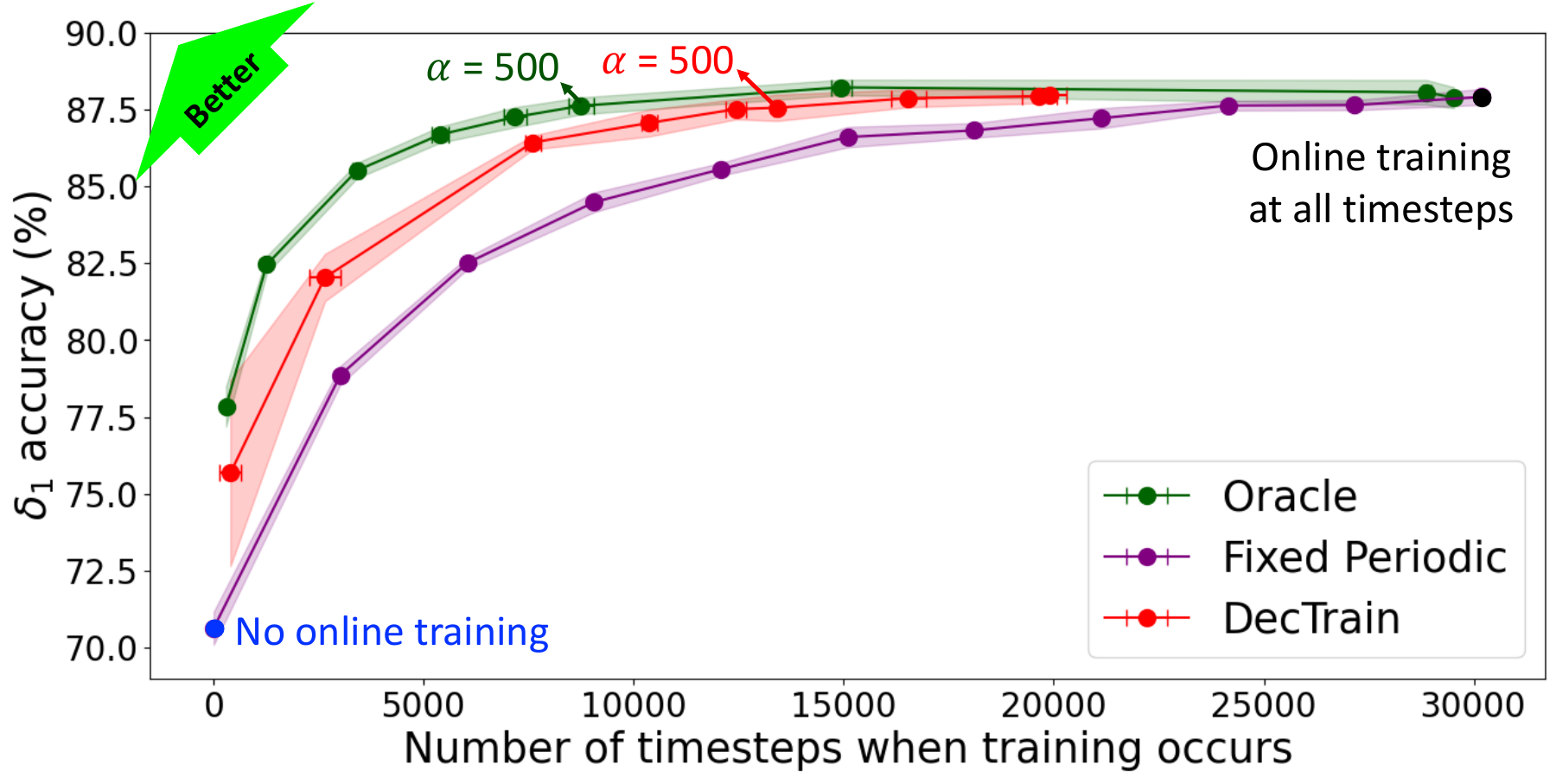}
    \caption{$\delta_1$ accuracy vs. number of timesteps when training occurs ($N_{train}$) on Exp. 5 for each method.
    Shaded regions and error bars represent one standard deviation of accuracy and $N_{train}$ respectively.
    Compared to fixed-periodic selection, DecTrain has a better trade-off on accuracy vs. $N_{train}$.}
    \label{fig:on_the_fly_learning_training_decisions_true_vs_acc}
\end{figure}
\subsection{Comparison to baseline selection methods}
We compare DecTrain to two baselines: (1) no online training, and (2) online training at all timesteps.
Table~\ref{table:comp_acc_all_sequences} summarizes the online training accuracy ($\delta_{1}$), validation accuracy ($\delta_{1,\text{val}}$), and GFLOPs on Exp. 1-10, where $\delta_{1}$ is the percentage of pixels within 25\% of the ground-truth. DecTrain maintains accuracy within 1\% of online training at all timesteps while reducing overall computation by 27\% on average as shown in Fig. \ref{fig:acc_gflops_barchart}. 
We also report the percentage of timesteps where DecTrain's decision is to train.
Given a fixed $\alpha=500$ set without prior knowledge of the online data, DecTrain can decide the percentage of timesteps to train specific to each indoor environment while maintaining accuracy. We also test DecTrain on an extremely out-of-distribution outdoor environment (Exp. 11) where the pre-trained monocular depth DNN only gets 0.1\% accuracy with no online training.
DecTrain achieves a 69.8\% $\delta_1$ accuracy by deciding to train 70\% of the time, showing DecTrain decides to train more in more out-of-distribution environments. There is a 4\% accuracy gap compared to online training at all timesteps which is larger than on indoor datasets ($<1$\%), due to the decision DNN needing time to adapt to the out-of-distribution environment.

\par Fig.~\ref{fig:qualitative_imgs} shows the qualitative results of the baselines and DecTrain ($\alpha=500$) on Exp. 5. 
DecTrain achieves similar depth prediction and error as online training at all timesteps, and both improve upon the no online training baseline. 
\subsection{Comparison to greedy oracle}
Fig.~\ref{fig:on_the_fly_learning_histogram} shows histograms of training decisions for (1) online training at all timesteps, (2) a greedy oracle as defined in Sec.~\ref{sec:dectrain} that uses the ground-truth utility of training, and (3) DecTrain. We see that both the greedy oracle and DecTrain reduce the timesteps when training occurs. Furthermore, DecTrain is making decisions similar to the decisions made by the oracle. However, a limitation of DecTrain is that the decision DNN tends to overestimate the utility of training.
\subsection{Comparison to fixed-periodic selection strategy}
Fig.~\ref{fig:on_the_fly_learning_training_decisions_true_vs_acc} shows accuracy vs. the number of timesteps when training occurs ($N_{train}$) for (1) the greedy oracle, (2) DecTrain, and (3) fixed-periodic selection on a multiple environment experiment (Exp. 5). 
For the greedy oracle and DecTrain, we sweep over $\alpha=50,100,200,300,400,500,1000,10^4, 10^9$.
For fixed-periodic selection, we sweep over 11 patterns that train on $\beta = 0\%,10\%,20\%,\cdots,100\%$ of the timesteps.
\par Using the greedy oracle achieves the best accuracy and $N_{train}$ trade-off, 
meaning that the ground-truth relative loss improvement is a good utility for deciding when to train.
However, we do not have access to ground-truth relative loss improvement when making a decision whether to train. 
Compared to fixed-periodic selection, DecTrain achieves higher accuracy at the same number of training steps by considering margin and ability to improve.
In addition, fixed-periodic selection results in a constant percent of the sequence trained regardless of the environment for a fixed $\beta$; 
meanwhile, in DecTrain, a fixed $\alpha$ that is set a priori
results in different amounts of training specialized for each environment.
\begin{figure}[t] \centering
     \centering
     \includegraphics[width=\columnwidth]{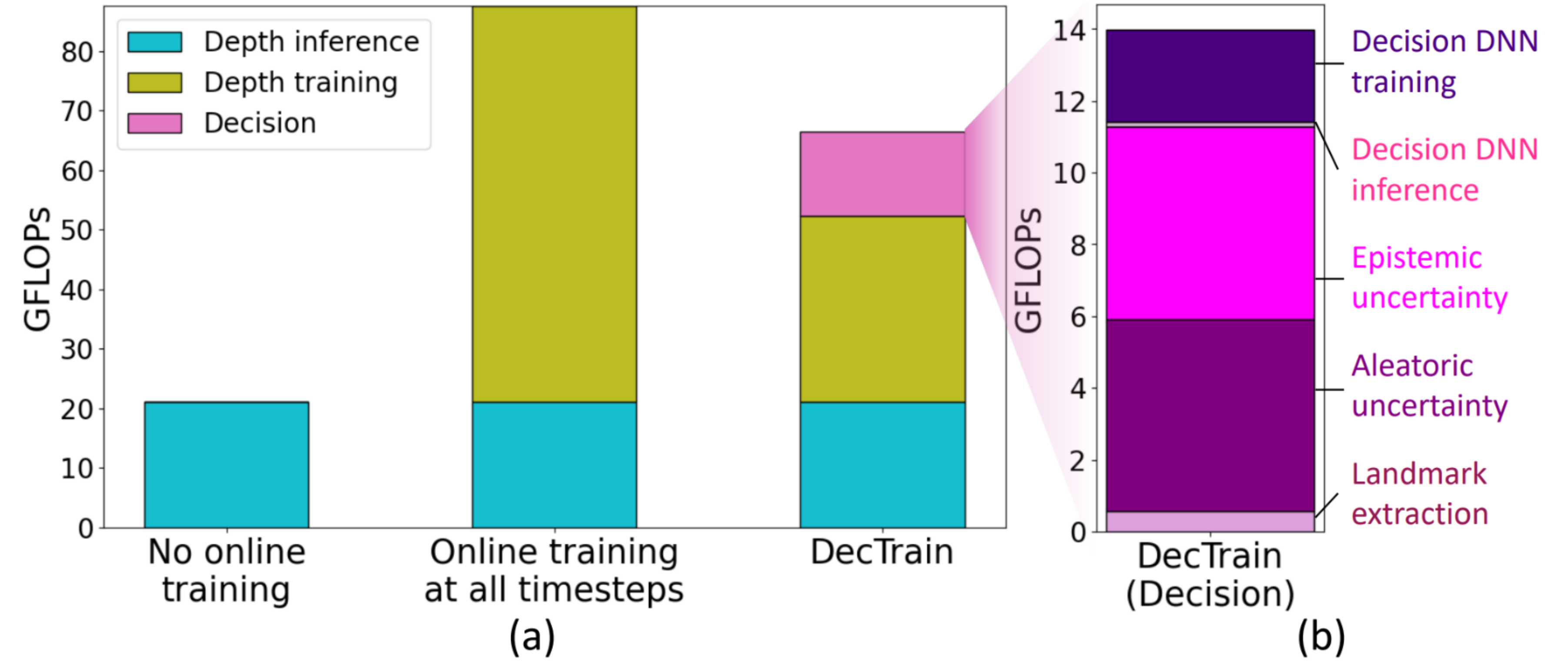}
     \caption{(a) Breakdown of the average compute per timestep of 
     no online training, online training at all timesteps, and DecTrain for Exp. 5.
     (b) Breakdown of DecTrain's decision overhead. Uncertainty estimation dominates the overhead.}
     \label{fig:FLOPs_breakdown}
\end{figure}

\begin{figure}[t]
\vspace{0.15cm}
    \centering
    \includegraphics[width=0.98\columnwidth]{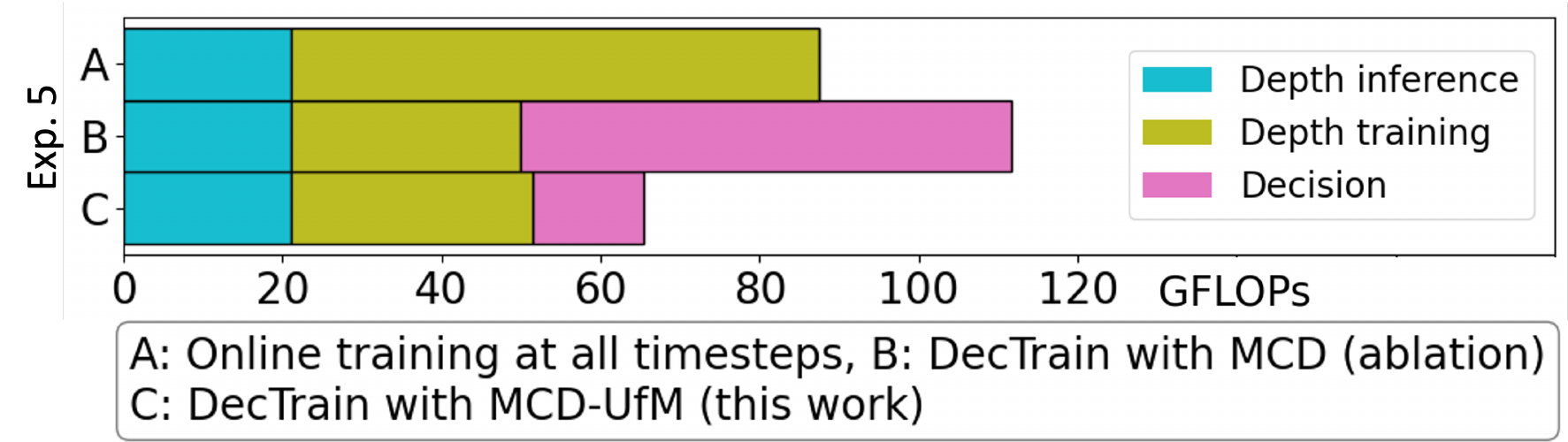}
    \caption{
    DecTrain with a lightweight epistemic uncertainty estimation (MCD-UfM~\cite{sudhakar2022uncertainty}) achieves compute savings, while using MCD does not due to the large overhead.
    }
    \label{fig:FLOPs_different_epistemic_unc}
\end{figure}

\begin{table}[ht]
\vspace{0.15cm}
\begin{center}
\caption{Ablation study for aleatoric and epistemic uncertainty showing KL divergence from oracle (lower is better).
}
\begin{tabular}{c | c c c c}
\toprule
   & \multicolumn{4}{c}{KL divergence [bits] ($\downarrow$) vs. greedy oracle} \\
 Exp. & DecTrain & No aleatoric & No epistemic & No A or E\\
 \toprule
 1 & \textbf{0.01} & 0.15 & 0.20 & 0.10 \\
 2 & \textbf{0.02} & 0.10 & 0.05 & 0.11 \\
 3 & \textbf{0.02} & 0.43 & 0.08 & 0.73 \\
 4 & \textbf{0.02} & 0.62 & 0.04 & 0.42 \\
 5 & \textbf{0.01} & 0.04 & 0.05 & 0.07 \\
 6 & \textbf{0.04} & 0.89 & 0.11 & 0.50 \\
 7 & \textbf{0.05} & 0.94 & 0.11 & 1.30 \\
 8 & \textbf{0.04} & 0.12 & 0.06 & 0.53 \\
 9 & \textbf{0.02} & 0.05 & 0.09 & 0.20 \\
 10 & 0.05 & 0.47 & \textbf{0.01} & 0.38 \\
\bottomrule
\end{tabular}
\label{table:ablation}
\end{center}
\vspace{-8pt}
\end{table}

\begin{figure}[t] \centering
        \begin{subfigure}[b]{0.95\columnwidth}
             \centering
             \includegraphics[width=\textwidth]{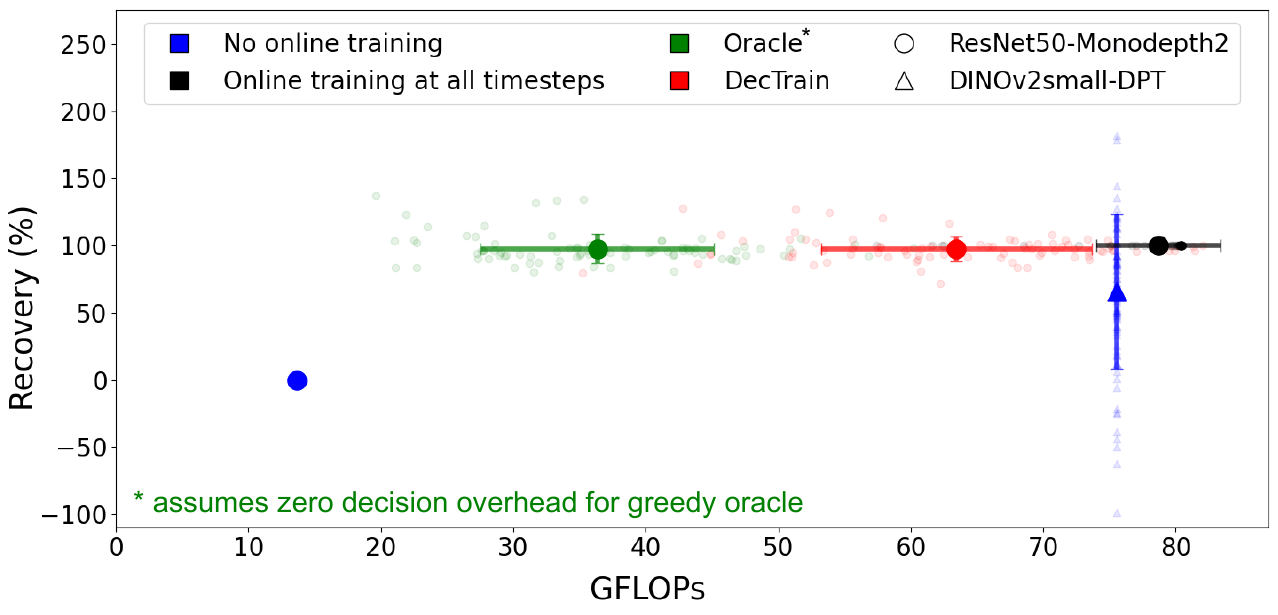}
         \end{subfigure}
         \caption{We show that online training a low-cost DNN with DecTrain (red $\iscircle$) can be more compute and accuracy efficient than a more generalizable high-cost DNN (blue $\triangle$).}
         \label{fig:low_cost_vs_high_cost_dnn}
\end{figure}
\subsection{Computational overhead}
Since DecTrain introduces components of the overhead that are either run at every timestep or only on timesteps when DecTrain decides to train, we modify Eq.~\ref{eq:compute_cost_formula} to be $C_{tot} = N \cdot C_{inf} + N_{train} \cdot C_{train} + N\cdot C_{dec_1} + N_{train} \cdot C_{dec_2}$ 
where $C_{inf}$ refers to the computation of the monocular depth DNN inference, $C_{train}$ refers to the computation of the online training process for the decoder of the monocular depth DNN, 
$C_{dec_1}$ refers to the decision overhead run at every timestep (computing inputs for decision DNN, decision DNN inference, computing greedy policy), and $C_{dec_2}$ refers to the decision overhead run at timesteps we train on (obtaining label for utility of training, training decision DNN). \par Fig.~\ref{fig:FLOPs_breakdown}(a) shows the compute breakdown of the average compute per timestep for the baselines and DecTrain
on Exp. 5.
We see that DecTrain achieves lower average compute per timestep compared to online training at all timesteps despite introducing 
overhead.
Fig.~\ref{fig:FLOPs_breakdown}(b) shows the compute breakdown of DecTrain decision overhead dominated by aleatoric and epistemic uncertainty estimation.
Despite the overhead, both aleatoric and epistemic uncertainty are important for DecTrain.
Table~\ref{table:ablation} shows the decision quality of DecTrain using four different sets of inputs to the decision DNN: (1) all mentioned in Section~\ref{subsec:utility_prediction} (this work), (2) all except aleatoric uncertainty statistics (ablation), (3) all except epistemic uncertainty statistics (ablation), and (4) all except aleatoric and epistemic uncertainty statistics (ablation).
We evaluate the decision quality by the KL divergence between the decisions made by the greedy oracle and the decisions made by each method, using the empirical probability density functions constructed from the decision histograms.
DecTrain achieves the lowest KL divergence in most experiments, showing that including both aleatoric and epistemic uncertainty obtains the closest decision to the greedy oracle.
\par To ensure the overhead does not negate the savings from reducing $N_{train}$, the choice of epistemic uncertainty algorithm is key.
Fig.~\ref{fig:FLOPs_different_epistemic_unc} shows the breakdown of the average compute per timestep of (1) online training at all
timesteps, (2) DecTrain with epistemic uncertainty from MC-Dropout (MCD)~\cite{gal2016dropout} (ablation), (3) DecTrain with epistemic 
uncertainty from MC-Dropout-UfM (MCD-UfM)~\cite{sudhakar2022uncertainty} (this work).
While MCD is widely used to estimate epistemic uncertainty, 
it causes the overhead to outweigh the savings from reducing $N_{train}$. Instead, using a more compute-efficient method such as MCD-UfM avoids negating the savings.
\subsection{Comparison to a SOTA zero-shot monocular depth DNN}
To show the impact of DecTrain on overall accuracy and computation trade-offs, we compare ResNet50-Monodepth2 with DecTrain (low $C_{inf}$, nonzero $N_{train}$) to DINOv2small-DPT (high $C_{inf}$, $N_{train}=0$).
DINOv2small-DPT is a SOTA monocular depth DNN knowledge distilled for resource-constrained zero-shot applications~\cite{oquab2023dinov2}; the $C_{inf}$ is 5.5$\times$ compared to $C_{inf}$ of ResNet50-Monodepth2.
We refer to DINOv2small-DPT as the high-cost DNN and ResNet50-Monodepth2 as the low-cost DNN for the rest of the paper. 
On average on 100 sequences of Exp. 12, the low-cost DNN with DecTrain has 6\% higher accuracy (81\%) compared to the high-cost DNN (75\%) while using 17\% less computation.
\par Fig.~\ref{fig:low_cost_vs_high_cost_dnn} shows the recovery vs. compute of the low-cost DNN with (1) no online training, (2) online training at all timesteps, (3) greedy oracle ($\alpha=500$), and (4) DecTrain ($\alpha=500$) as well as (5)
the high-cost DNN with no online training on Exp. 12.
Recovery $\tau$ is the relative accuracy gain recovered by the method compared to the low-cost DNN with online training at all timesteps as in
\begin{equation}
    \tau = \frac{\delta_{1,\text{method}}-\delta_{1,\text{no\_online\_training}}}{\delta_{1,\text{online\_training\_at\_all\_timesteps}}-\delta_{1,\text{no\_online\_training}}}\times100\%.
\end{equation}
We show recovery results on sequences where online training helps significantly (online training at all timesteps improves $>5$\% accuracy over no online training),
which
are 84 out of the 100 sequences. 
While the low-cost DNN with online training at all timesteps has 100\% recovery, it is also the most computationally expensive method. 
Training the low-cost DNN with DecTrain achieves 97\% recovery while reducing computation by 16\% on average compared to the high-cost DNN which achieves only 66\% recovery. 
The greedy oracle performs best in achieving the same recovery as DecTrain at 97\% with an assumed zero overhead, providing an upper bound on the accuracy and computation trade-off.
On a few individual sequences where DecTrain decides to train most of the time in Fig.~\ref{fig:low_cost_vs_high_cost_dnn}, we see that the decision-making overhead causes the overall compute to be higher than online training at all timesteps, though on average, DecTrain makes significant savings.
Moreover, on the 16 sequences where online training does not improve or even hurts accuracy, online training at all timesteps reduces accuracy by 5\% compared to no online training, while DecTrain only reduces accuracy by 1\%, showing that DecTrain can intelligently decide when training is harmful and choose to train less. 
\par These results also extend to an even smaller DNN (ResNet18-Monodepth2) with 12$\times$ lower inference cost compared to the high-cost DNN.
On 85 sequences where online training helps significantly, 
using this smaller DNN with DecTrain achieves a better recovery (89\%) while reducing computation by 56\% on average compared to the high-cost DNN (77\% recovery). 
On all 100 sequences, the average accuracy is 4\% higher than the high-cost DNN while reducing computation by 57\%.
Thus, we show that a smaller DNN with DecTrain enables a better accuracy and computation trade-off compared to a larger, more generalizable DNN. 
%%%%%%%%%%%%%%%%%%%%%%%%%%%%%%%%%%%%%%%
\section{Conclusion}
In this work, we introduce DecTrain, a new algorithm that decides when to train a monocular depth DNN online. 
Showing competitive accuracy and computational savings compared to performing online training at all timesteps, DecTrain highlights the importance of deciding wisely and efficiently to deliver a higher accuracy at lower cost when adapting to a new environment.
%%%%%%%%%%%
% elaboration on limitation
%%%%%%%%%%%
%\hl{DecTrain is limited when deployed to environments which requires frequent online training to increase or maintain accuracy. Specifically, in extremely out-of-distribution environments, DecTrain encounters larger accuracy loss due to the choice of utility function and incorrect decisions made while the decision DNN is adapting.
%Additionally, the decision-making overhead can outweigh the savings of reducing the cost of online training when DecTrain decides to train most of the time.}
%%%%%%%%%%%
Future work includes incorporating different utility functions and non-myopic decision-making,
% reducing the frequency of the decision-making to reduce the overhead, 
% reducing the overhead,
% for environments requiring frequent online training, 
as well as testing in real-world robotic deployments.
%\hl{Future work includes testing DecTrain in real-world robotic deployments and improving the utility function for non-myopic decision-making.}

% The impact of this work shows that a smaller DNN that is less generalizable that we can spend energy running online training can be more efficient at similar accuracy than a 
% make references section 
% \FloatBarrier
\bibliographystyle{IEEEtran}
\bibliography{IEEEabrv,bib}

\end{document}